\documentclass[runningheads]{llncs}
\usepackage{graphicx}
\usepackage{comment}
\usepackage{amsmath,amssymb} % define this before the line numbering.
\usepackage{color}
\usepackage[pagebackref=true,breaklinks=true,letterpaper=true,colorlinks,bookmarks=false]{hyperref}

\usepackage[width=122mm,left=12mm,paperwidth=146mm,height=193mm,top=12mm,paperheight=217mm]{geometry}

\begin{document}
% \renewcommand\thelinenumber{\color[rgb]{0.2,0.5,0.8}\normalfont\sffamily\scriptsize\arabic{linenumber}\color[rgb]{0,0,0}}
% \renewcommand\makeLineNumber {\hss\thelinenumber\ \hspace{6mm} \rlap{\hskip\textwidth\ \hspace{6.5mm}\thelinenumber}}
% \linenumbers
\pagestyle{headings}
\mainmatter

\title{Augmented Parallel-Pyramid Net for Attention Guided Pose-Estimation} % Replace with your title

% INITIAL SUBMISSION 
%\begin{comment}
%\titlerunning{Augmented Parallel-Pyramid Net for Attention Guided Pose-Estimation} 
%\authorrunning{ECCV-20 submission ID \ECCVSubNumber} 
%\author{Luanxuan Hou}
%\institute{Paper ID \ECCVSubNumber}
%\end{comment}
%******************

% CAMERA READY SUBMISSION
%\begin{comment}
\titlerunning{Augmented Parallel-Pyramid Net for Attention Guided Pose-Estimation}
% If the paper title is too long for the running head, you can set
% an abbreviated paper title here
%
\author{Luanxuan Hou\inst{123} \and
Jie Cao\inst{1} \and
Yuan Zhao\inst{3} \and
Haifeng Shen\inst{3} \and
Yiping Meng\inst{3} \and
Ran He\inst{1} \and
Jieping Ye\inst{3}}
\authorrunning{L. Hou et al.}
% First names are abbreviated in the running head.
% If there are more than two authors, 'et al.' is used.
%
\institute{Center for Research on Intelligent Perception and Computing (CRIPAC), Institute of Automation, Chinese Academy of Sciences, Beijing, China\\\email{\{luanxuan.hou, jie.cao\}@cripac.ia.ac.cn}, \email{ran.he@ia.ac.cn} \and
School of Artificial Intelligence, UCAS, Beijing, China
\\ \and
AI Tech, Didi Chuxing, Beijing, China\\
\email{\{zhaoyuanjason, shenhaifeng, mengyipingkitty, yejieping\}@didiglobal.com}}
%\end{comment}
%******************
\maketitle

\begin{abstract}
The target of human pose estimation is to determine body part or joint locations of each person from an image. This is a challenging problems with wide applications. To address this issue, this paper proposes an augmented parallel-pyramid net with attention partial module and differentiable auto-data augmentation. Technically, a parallel pyramid structure is proposed to compensate the loss of information. We take the design of parallel structure for reverse compensation. Meanwhile, the overall computational complexity does not increase. We further define an Attention Partial Module (APM) operator to extract weighted features from different scale feature maps generated by the parallel pyramid structure. Compared with refining through upsampling operator, APM can better capture the relationship between channels. At last, we proposed a differentiable auto data augmentation method to further improve estimation accuracy. We define a new pose search space where the sequences of data augmentations are formulated as a trainable and operational CNN component. Experiments corroborate the effectiveness of our proposed method. Notably, our method achieves the top-1 accuracy on the challenging COCO keypoint benchmark and the state-of-the-art results on the MPII datasets. 

%Human pose-estimation aims to determine body part or joint positions of each person from an image. Since some in-the-wild person positions are often difficult to predict (resulting in some hard keypoints), it is still challenging for pose estimation methods to achieve high performance. To address this issue, this paper proposes an augmented parallel-pyramid net with attention partial module and differentiable auto-data augmentation. First, a parallel pyramid structure is proposed to compensate for the loss of information. We take the design of parallel structure for reverse compensation. At the same time, the computational complexity does not increase. Second, we define an Attention Partial Module (APM) operator to extract weighted features from different scale feature maps generated by the parallel pyramid structure. Compared with refining through upsampling operator, APM can better capture the relationship between channels. Lastly, we proposed a differentiable auto data augmentation method to further improve estimation accuracy. We define a new pose search space where the sequences of data augmentations are formulated as a trainable and operational CNN component. Experiments show that our network achieves the top-1 accuracy on the challenging COCO keypoint benchmark and the state-of-the-art results on the MPII datasets.
\end{abstract}

\section{Introduction}

Multi-person pose-estimation has been intensely investigated in computer vision due to its wide applications. There are many challenges in multi-person pose-estimation, such as occluded, self-occluded and invisible keypoints. For example, the visibility of keypoints is greatly affected by wearing, posture, viewing angle and backgrounds. With the advances of deep learning, there are growing interests in developing deep neural networks for multi-person pose-estimation \cite{fang2017rmpe,cao2017realtime,kocabas2018multiposenet,sun2019deep,papandreou2018personlab}.
%-------------------------------------------------------------------------
\begin{figure}[htb]
\begin{center}
\includegraphics[width=1\linewidth]{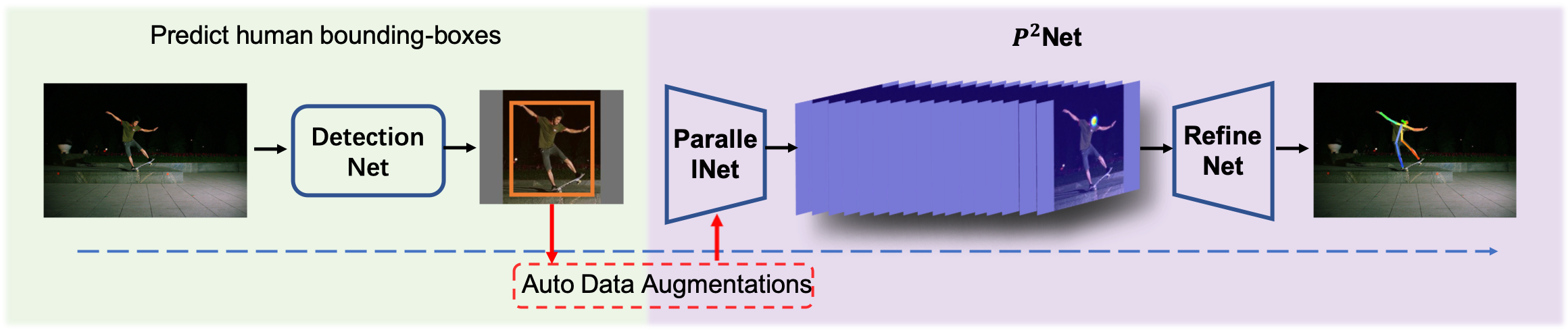}
\end{center}
   \caption{The image is fed into a detector to generate the bounding boxes and then the detected person is fed into the $P^2$ Net to generate the heatmap of the $N$ channel. The maximum value of each heatmap $(x,y)$ indicates the position of the detection point.}
\label{fig:1}
\end{figure}
%-------------------------------------------------------------------------
Multi-person pose-estimation models can be categorized into top-down and bottom-up methods. As shown in Figure~\ref{fig:1}, the top-down methods firstly detect each person by some algorithms (e.g., FPN \cite{szegedy2017inception}, Mask RCNN \cite{he2017mask}, TridentNet \cite{li2019scale}), and then generate keypoints in these bounding-boxes. The representative methods include Simple-Baseline \cite{xiao2018simple}, RMPE \cite{fang2017rmpe}, CPN \cite{chen2018cascaded} and HRNet \cite{sun2019deep}. Whereas, the bottom-up methods also consist of keypoints detection and clustering. They firstly detect all keypoints in a partial image, and then cluster all keypoints into different individuals. PAF \cite{cao2017realtime}, Associative Embedding \cite{newell2017associative}, Part Segmentation \cite{xia2017joint}, Mid-Range offsets \cite{papandreou2018personlab} are representative methods.

Although great progress has been made, there are still many challenges for accurate pose estimation in the wild. In the unconstrained conditions, the visibility of keypoints is greatly affected by wearing, posture, viewing angle, and backgrounds. Besides, the large pose variations further increase the difficulty in detection. Particularly, when there are overlapping parts of human body, the occluded keypoints are extremely difficult to detect. To increase the diversity of training data, we usually use data augmentation strategies. Reasonable data enhancement improves the robustness of the model. On the contrary, irrational setting of data enhancement parameters will also bring noise to the network. Moreover, data augmentation plays an important role in pose estimation. Existing augmentation strategies are usually hand-crafted or based on tuning techniques \cite{he2016deep,hu2018squeeze}. Due to the complexity of multi-person pose estimation, manual designing the data augmentation sequences is a non-trivial task. Hence, finding an efficient strategy becomes urgent.

FPN \cite{lin2017feature} exploits the inherent multi-scale, pyramidal hierarchy of deep convolutional networks to construct feature pyramids with marginal extra cost. A top-down architecture with lateral connections is developed for building high-level semantic feature maps at all scales. Due to the down-sampling of the top-down structure, we get more semantic information at different scales in the deep layer. But the spatial resolution of the deep feature map becomes lower, resulting in loss of spatial information.

To address these problems, we propose a novel two-stage network structure which consists of ParallelNet and RefineNet. ParallelNet adopts parallel structures to learn high-resolution representations to compensate for the information loss. On the one hand, our ParallelNet extracts sufficient context information necessary for the inference of the occluded and invisible keipoints. On the other hand, it effectively preserves the spatial information and the semantic information of the pyramid feature network. Based on the pyramid features, our RefineNet explicitly addresses the hard keypoint detection problem by optimizing an online hard keypoints mining loss \cite{shrivastava2016training}. The general refining operation is to perform upsampling and then concatenation, which ignores the relationships between feature maps with different scales. Alternatively, we maintain the spatial resolution of the features by employing a dilated bottleneck structure. In order to make RefineNet focus on more informative regions, we add an Attention Partial Module (APM) to the output of ParallelNet. Our work not only maintains high-resolution feature maps but also keeps large receptive field, both of which are important for pose detection and estimation. 

We address the multi-person pose-estimation problems based on a top-down pipeline. Human detector is firstly adopted to generate a set of human bounding-boxes, followed by our network for keypoints localization in each human bounding-box. In addition, we also explore the effects of various factors that might affect the performance of multi-person pose-estimation, including person detector. 

We propose an automated approach (named Auto-Pose) to search for data augmentation strategies instead of hand-crafted data augmentation strategies. Our approach is explicitly suited to the pose-estimation task. Auto-Pose designs  a new pose search space to encode the sequences of data augmentation that is commonly used in pose-estimation. Moreover, it equips with the differentiable method instead of reinforcement learning, making the searching results particularly suitable for pose-estimation. We summarize our contributions as follows:
\begin{enumerate}
    \item \textbf{First}, we improve the upper part of the pyramid structure, retaining both the global and the local information effectively. We propose a repeated multi-scale fusions operation and acquire high-resolution representations to compensate for the information losses incurred by the single pyramid structure. Furthermore, we implement Attention Partial Module (APM) to probe the relationships between feature maps with different scales. Combining these techniques mentioned above, the predicted keypoints heatmap and spatial position can be more accurate.
    \item \textbf{Second}, to the best of our knowledge, we are the very first to search the sequence of data augmentations for human pose-estimation, replacing the laborious manual design of data augmentation. We propose a novel pose search space where the sequences are formulated as a trainable and operational CNN component.
\end{enumerate}

%-------------------------------------------------------------------------
\begin{figure*}[htb]
    \centering
    \includegraphics[scale=0.24]{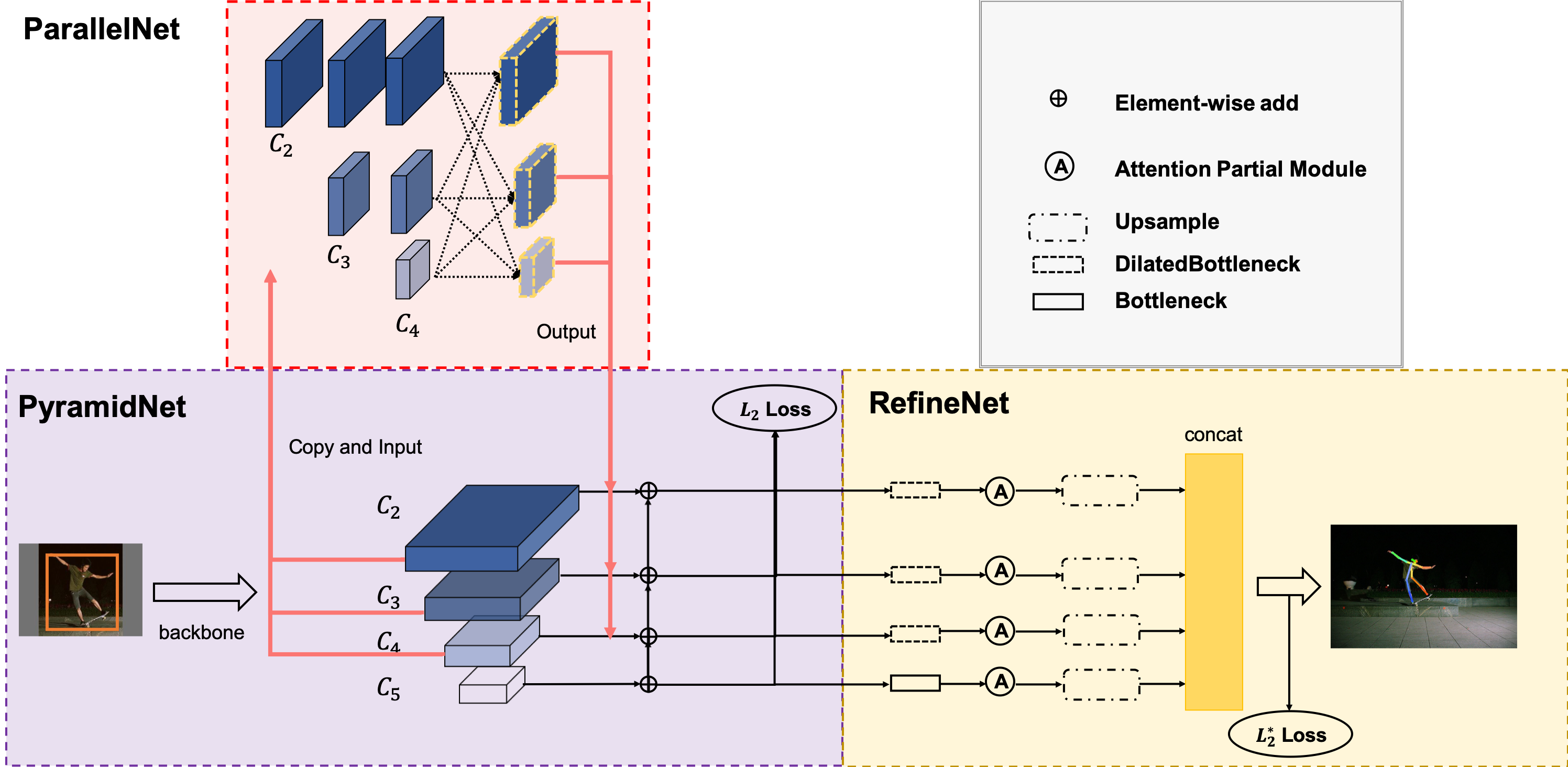}
    \caption{The architecture of the proposed pose-estimation network. The network adopts the idea of cascading pyramid network, which is divided into three sub-networks, PyramidNet, ParallelNet and RefineNet. The PyramidNet part(purple area) is a pyramid structure, and the ParallelNet is a parallel network structure(red area). In ParallelNet, we perform repeated multi-scale fusions, using parallel structures, resulting in high resolution representations to maintain the information in the pyramid network completely. Therefore, the predicted keypoints heatmap may be more accurate and the space is more accurate. In RefineNet(yellow area), correcting keypoints that are difficult to locate in ParallelNet. Before element-wise sum process, we adopt 1$\times$1 conv. In order to ensure the same number of channels.}
    \label{fig.2}
\end{figure*}
%-------------------------------------------------------------------------
\section{Related Works}
Human pose-estimation has been widely used in action recognition, human body structure generation, and etc. With the development of Convolutional Neural Networks (CNN), the mainstream methods have evolved from the HOG \cite{dalal2005histograms} and deformable parts model rely on handcraft features and graphical models to CNN that significantly improves the performance of pose-estimation. In general, there are two mainstream approaches: Top-Down and Bottom-Up.

{\bf Top-Down Methods}
Top-Down Methods are more like the human perception of generating keypoints after a person sees a human body. These methods can be categorized into two processes. Firstly, some detection networks such as FPN \cite{lin2017feature}, YOLO \cite{redmon2016you}, SSD \cite{liu2016ssd} are used to detect the person from the image and predict human bounding-boxes. Then in the human bounding-box, keypoints are predicted. Recent pose-estimation methods apply object detection for making predictions. The U-shape stacked hourglass network \cite{newell2016stacked} is often used to stack up several hourglass modules to generate prediction. Zisserman applies some RNN like architectures to sequentially refine the results \cite{belagiannis2017recurrent} . These methods depend on the reliability of the person detector so that they often face difficulty in recovering the poses that are obscured or difficult to locate. Thus, the output predictions of top-down methods might potentially benefit from an additional refinement step.

{\bf Bottom-Up Methods}
Bottom-up methods firstly predict all body joints and then group them into full poses. Instead of applying person detection, these methods rely on context information and inter body joint relationships, therefore they are usually used in real-time scenarios. Pishchulin\cite{insafutdinov2016deepercut} proposed a bottom-up approach that jointly labeled part detection candidates and associated them to individual people. Zhe\cite{cao2017realtime} maped the relationships between keypoints and assembled detected keypoints into different poses of people. Newell\cite{newell2017associative} simultaneously produced score maps and pixel-wise embedding to group the candidate keypoints to different people to get final multi-person pose-estimation. However, it is often difficult to model the joint relationships, resulting in some failure cases to disambiguate pose of different people or group body parts of the same person into different clusters.

{\bf Auto Data Augmentations}
Recently, some researchers adopted AutoAugment search space with improved optimization algorithms to seek more efficient policies \cite{ho2019population,lim2019fast}. Most of AutoML approaches search CNN on a small proxy task and automatically tune hyperparameters. Our work is motivated by recent researches on AutoML \cite{cubuk2018autoaugment} and Neural Architecture Searching (NAS) \cite{baker2016designing,real2019regularized,zoph2018learning}, while we focus on searching for a pose estimation model with high performance. Inspired by the efficiency of NAS and AutoML algorithms, we develop new auto methods for the pose-estimation problem. Particularly, we focus on the searching differentiable sequences of data augmentations in network. Experimental results demonstrate that sequence information affects estimation results and validate that auto strategy significantly outperforms simply adopting the hand-crafted sequences of data augmentations based on experience of engineering.

\section{Proposed Method}

We introduce the proposed top-down pose estimation method. As shown in Figure~\ref{fig.2}, our $P^2$ Net involves three sub-networks: ParallelNet, PyramidNet and RefineNet. A human detector is firstly applied on an image to generate a set of human bounding-boxes, and then the keypoints for each person are located by our proposed pose-estimation network, \textbf{$P^2$ Net}.

\subsection{Human Detector}

FPN \cite{lin2017feature} uses a pyramid structure to maintain the balance of spatial resolution and semantic information. In FPN, a large object is generated and predicted within deeper layers. Since FPN adopts the top-down method like the inverted pyramid structure, the boundary of these objects may be too blurry to get an accurate regression. FPN predicts small object in shallower layers. However, shallow layers only have low semantic information so that there may be not sufficient to recognize the category of the object instances. Therefore, detectors must enhance their classification capability by involving context cues of high-level representations from the deeper layers. Hence, FPN is often enhanced by adopting a bottom-up pathway. However, the missing of small objects in deeper layers still leads to losing context cues.

To address these problems, we use the feature activation output by the last residual block of each stage in ResNet. We denote the output of conv \textit{i} as \textit{$P_i$} (i=2,3,4,5). The detector has exactly the same number of stages as the used detector such as FPN where we keep the scale of \textit{$P_4$} ,\textit{$P_5$}, and \textit{$P_6$} the same. We keep the spatial resolution 16$\times$downsample after \textit{$P_4$}. Then, we apply bottleneck with dilation \cite{yu2015multi} as a basic network block to maintain the receptive filed efficiently. Since dilated convolution is still time consuming, \textit{$P_5$} and \textit{$P_6$} keep the same channels as \textit{$P_6$} setting \textit{C}=256.In Section~\ref{exp} we also discuss the impact of different detection networks on pose estimation.

\subsection{Pyramid Network With Information Compensation}

\begin{figure}[htb]
    \centering
    \includegraphics[scale=0.2]{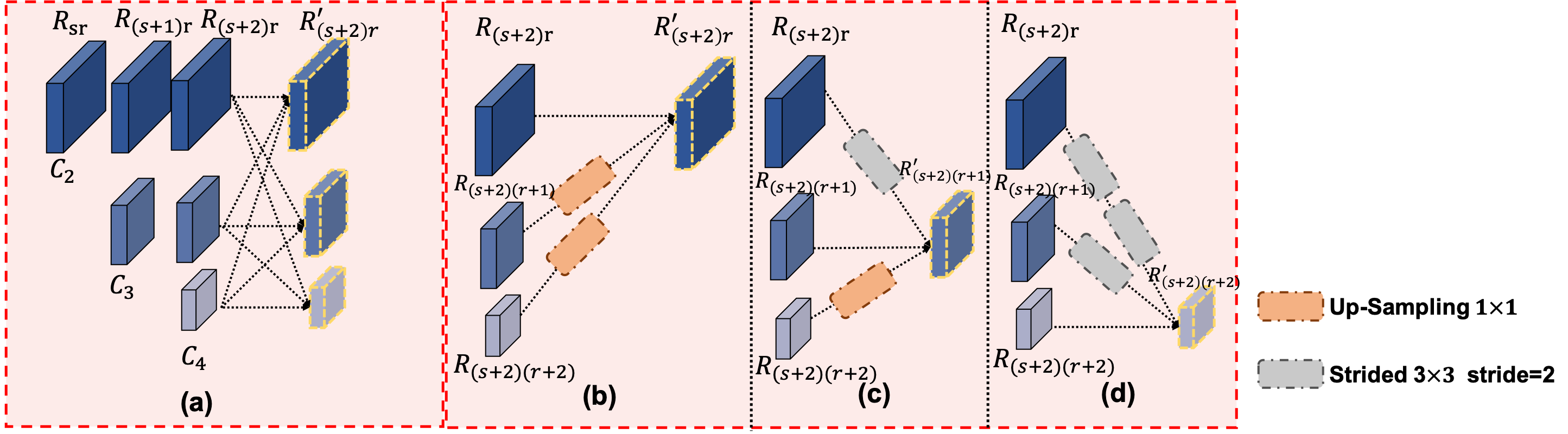}
    \caption{(a) represents the overall architecture of the proposed parallel structures. Before element-wise sum process, we adopt 1$\times$1 conv. In order to ensure the same number of channels. (b)(c)(d) represent how the exchange unit aggregates the information for high, medium and low resolutions from the left to the right, respectively.}
    \label{fig.3}
\end{figure}

\textbf{ParallelNet - Making up for information loss}. We design our $P^2$ Net based on the backbone of ResNet. We denote the output of these last residual blocks as, \textit{$C_r$} (r=2,3,4,5) for conv2, conv3, conv4, and conv5 feature map outputs respectively. Convolution filters (kernel size$=$3$\times$3) are applied on \textit{$C_r$} (r=2,3,4,5) to generate the heatmaps of keypoints. As we know, the shallow features have the high spatial resolution for localization but low semantic information for recognition. Deep feature layers have more semantic information but spatial resolution is not high enough. Therefore, we apply the pyramid structure in PyramidNet. The advantage of using the pyramid structure is that pyramid structure can effectively preserve the semantic information and spatial resolution of the feature maps. The scale of the upper layer ${C_r}$ is twice that of the next layer ${C_{r+1}}$. We use the connection method depicted in Figure~\ref{fig.3}. Then we use $R_r$ to represent the spatial resolution of ${C_r}$.The resolution has the following calculation:
\begin{equation}
   R_r = 2R_{r+1}.
\end{equation}

In parallel structure, as shown in Figure~\ref{fig.3}, where ${R_{sr}}$ represents the feature map of each stage \textit{Sth} and \textit{r} is the index of resolution.

Existing networks for pose estimation are built by connecting high  representation to low representation subnetworks in series, where each subnetwork, forming a stage, is composed of a sequence of convolutions and there is a down-sample layer across adjacent subnetworks to halve the resolution. But we keep the same spatial resolution of each subnetworks to ensure sufficient feature expression:
\begin{equation}
   R_{sr} = R_{(s+1)r} = R_{(s+2)r}. 
\end{equation}

 ${R}'_{_{(s+i)r}}$ represents the result of fusion of various scales through element-sum. As shown in Figure 3 (b-d), the operation in (c) can be interpreted as:
 \begin{equation}
   {R}'_{_{(s+2)(r+1)}} = O(R_{(s+2)r}) + R_{(s+2)(r+1)} + {O}'(R_{(s+2)(r+2)}). 
\end{equation}
 
 $O$ represents one strided 3$\times$3 convolution and ${O}'$ represents 
 the operation of upsampling.
 
 One strided 3$\times$3 convolution with the stride=2 for 2$\times$downsampling, and two consecutive strided 3$\times$3 convolutions with the stride=2 for 4$\times$ downsampling. For upsampling, we adopt the simple nearest neighbor sampling following a 1$\times$1 convolution for aligning the number of channels. And we use one strided 3$\times$3 convolution with the stride = 2 for 2$\times$downsampling to complete the transformation.
\begin{figure}[htb]
    \centering
    \includegraphics[scale=0.2]{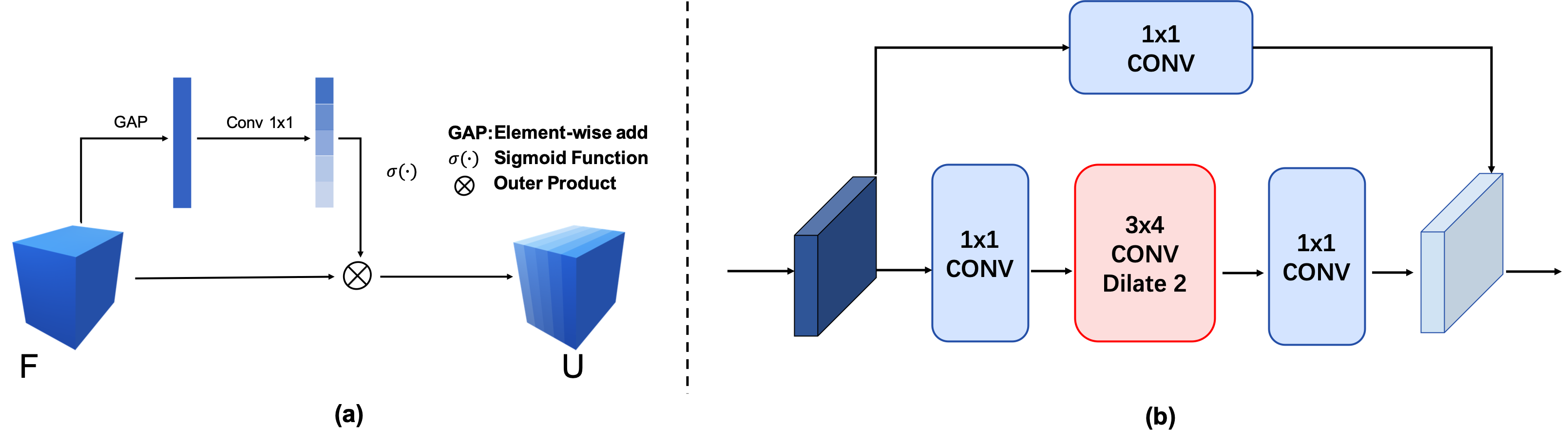}
    \caption{(a) Attention Partial Module. (b) Dilated Bottleneck with DilatedConv.}
    \label{fig.4}
\end{figure}

\subsection{Dilated Attention Partial Module}

\textbf{RefineNet - Better capture the re-lationship between channels}. Based on the feature pyramid representations generated by ParallelNet, we attach a RefineNet to refine the keypoints. In order to improve efficiency, our RefineNet transmits the information across different levels and finally integrates the information of different levels via upsampling and concatenating. Our RefineNet concatenates all the pyramid features rather than simply using the upsampled features at the end of hourglass module. 

In order to gather features selectivity from the shallow layer and the deep layer, we design the attention module, Attention Particial Module (APM). As shown in Figure~\ref{fig.4}(a), we denote the input feature maps as $F=\lbrack F_1,\dots F_c\rbrack\in\mathbb{R}^{N\times H\times W}$, then applying global average pooling to have output $T\in\mathbb{R}^{N\times1\times1}$, where $N$ denotes the number of channels. The $K$-th channel can be expressed as:
\begin{equation}
    T_k=\frac1{W\times H}\sum_i^H\sum_j^W(F_k(i,j)).
\end{equation}
$T$ is reorganized by a 1$\times$1 convolution layer $\phi$ with the same number of channels as $T$. The Sigmoid function $\sigma$ is applied to activate the convolution result, constraining the value of weight vector $V$. Then performing an outer product for $F$ and $V$, and the final output $U\in\mathbb{R}^{N\times H\times W}$ can be expressed as :
\begin{equation}
    U=F\otimes\sigma\lbrack\phi(T)\rbrack.
\end{equation}

In addition, we stack more Dilated bottleneck blocks as shown in Figure~\ref{fig.4}(b) into deeper layers, which can use smaller spatial-size conv to achieve a good trade-off between receptive filed and efficiency.

\subsection{Differentiable Sequences for Pose Data Augmentations}

\begin{figure}[htb]
    \centering
    \includegraphics[scale=0.13]{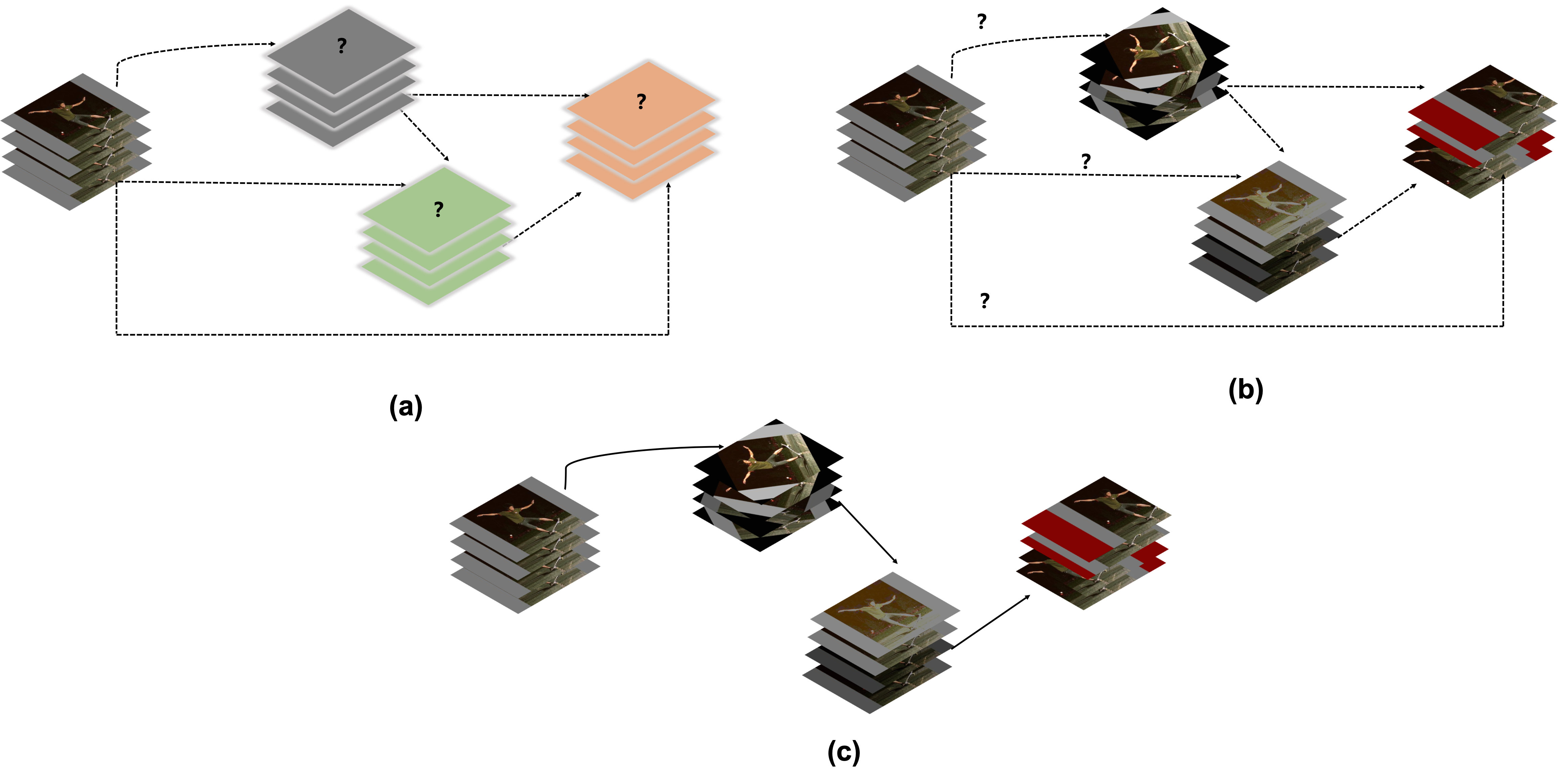}
    \caption{Four examples of learned sub-policies applied to one example image batch. (a) Each different color image batch corresponds to a different random sample of the corresponding sub-policy. Each step of an augmentation sub-policy consists of a triplet corresponding to the operation, the probability of application and a magnitude measure. (b) Operations on the edges are initially unknown. We take a Continuous relaxation of the search space by placing a mixture of candidate operations on each edge. We jointly optimize the mixed probability and network weights by solving the bi-level optimization problem. (c) We induce the example architecture from the learned mixing probabilities.}
    \label{fig.5}
\end{figure}

\textbf{More efficient}. Relying on engineering experience to manually set data augmentations is not innovative enough. Since our $P^2$ Net input is a cropped bounding-box image, we use these methods such as geometric transformation, color transformation, scale  transformation, etc. to increase data diversity. We treat data augmentation search as a discrete optimization problem. Based on previous work, in our search space, we define an augmentation policy as a unordered set of \textit{K} sub-policies. During training one of the \textit{K} sub-policies will be selected at random and then applied to the current image-batch. Each sub-policy has \textit{N} image transformations which we define 9 operations, as shown in Table~\ref{tab.1}.

\begin{table*}[htb]
    \caption{Table of the possible transformations that can be applied to an image. These are the transformations that are available to the controller during the search process.}
    \newcommand{\tabincell}[2]{\begin{tabular}{@{}#1@{}}#2\end{tabular}}
    \renewcommand{\arraystretch}{1.2}
    \setlength{\tabcolsep}{2mm}
    \centering
    \begin{tabular}{lp{9cm}l}
    \hline
    Operation Name  &   Description\\
    \hline
    TranslateX(Y)   &Translate the image and the bounding boxes in the horizontal (vertical) direction by $magnitude$ number of pixels.\\
    Rotate    &Rotate the image and the bounding boxes $magnitude$ degrees.\\
    Equalize  &Equalize the image histogram.\\
    Solarize   &Invert all pixels above a threshold value of $magnitude$.\\
    SolarizeAdd &For each pixel in the image that is less than 128, add an additional amount to it decided by the $magnitude$.\\
    Brightness&Adjust the brightness of the image. A $magnitude$=0 gives a black image, whereas $magnitude$=1 gives the original image.\\
    Sharpness&Adjust the sharpness of the image. A $magnitude$=0 gives a blurred image, whereas $magnitude$=1 gives the original image.\\
    Cutout&Set random square patches of pixels with side-length to gray.\\
    Scale&Scale with this $magnitude$.\\
    \hline
    \end{tabular}
    \label{tab.1}
\end{table*}

As shown in Figure~\ref{fig.5}, we turn this problem of searching for a learned augmentation policy into a discrete optimization problem by creating a search space. The search space consists \textit{K} = 3 sub-policies with each sub policy consisting of \textit{N} = 2 operations applied in sequence to a single image. At the same time, we set two hyperparameters, probability \textit{P} and magnitudes \textit{M} of operations. The probability parameter introduces a notion of stochasticity into the augmentation policy whereby the selected augmentation operation will be applied to the image with the specified probability. Because the range of \textit{M} for each data augmentations is different such as the range of random-scale (0.7$\sim$1.35) and the range of rotation ($-45^{\circ}$$\sim$$45^{\circ}$), we normalize \textit{M} to 0$\sim$10. Many existing methods for addressing the discrete optimization problem include reinforcement learning , evolutionary methods and sequential model-based optimization. In this paper, we apply a differentiable search algorithm. Let $\mathrm{O}$ be a set of candidate data augmentations where each operation represents function $o(\cdot)$ to be applied $x^i$, we relax the categorical choice of a particular operation to a softmax over all possible operations:
\begin{equation}
    \overline{o^{(i,j)}}(x)=\sum_{o\in\mathrm O}\frac{exp(\alpha_o^{(i,j)})}{\sum_{o'\in\mathrm O}exp(\alpha_{o'}^{(i,j)})}o(x).
\end{equation}
where the operation mixing weights for a pair of image batches $(i,j)$ are parameterized by a vector $\alpha$ of dimension $\mathrm O$. The task of sequences search then reduces to learning a set of continuous variables $\alpha=\{\alpha^{(i,j)}\}$. At the end of search, a discrete sequence of data augmentations can be obtained by replacing each mixed operation with the most likely operation. Let $\mathcal{L}_{train}$ and $\mathcal{L}_{val}$ denote the training loss and the validation loss, respectively. Both losses are determined by the data augmentation sequences architecture $\alpha$ and the weights $w$ in the network. The goal becomes the two formulas of alternating optimization:
 \begin{equation}
      \underset\alpha{min}\mathcal{L}_{val}(\omega^\ast(\alpha),\alpha)\;;\;\omega^\ast(\alpha)=argmin_\omega\mathcal{L}_{train}(\omega,\alpha).
 \end{equation}

 We adopt an approximation scheme as follows:
 \begin{equation}
    \nabla_\alpha\mathcal{L}_{val}(\omega^\ast,\alpha)\approx\nabla_\alpha\mathcal{L}_{val}(\omega-\zeta\nabla_\omega\mathcal{L}_{train}(\omega,\alpha),\alpha).
 \end{equation}
 where $\omega$ denotes the current weights maintained by the algorithm, and $\zeta$ is the learning rate for a step of inner optimization. %The sequences of data augmentations that are searched out is shown in Table~\ref{tab.2}.

%\begin{table}[!t]
%\caption{ The sub-policies used in our learned augmentation policy. For each image in each batch, one of the sub-policies is picked uniformly at random.}
%\renewcommand{\arraystretch}{1.3}
%\centering
%\setlength{\tabcolsep}{0.8mm}{
    %\begin{tabular}{ccccccc}
    %\hline
    %&Operation 1&P&M& Operation 2 &P&M\\
    %\hline
    %\textbf{Sub1}&BBox$\_$TranslateY&0.4&4&Cutout&0.6&8\\
    %\textbf{Sub2}&Rotate&0.5&10&Scale&0.8&6\\
    %\textbf{Sub3}&TranslateX&0.8&4&Equalize&0.8&10\\
    %\hline
    %\end{tabular}}
    %\label{tab.2}
%\end{table}

\section{Experiments and Analysis}
\label{exp}
We evaluate our proposed $P^2$ Net on two pose-estimation datasets. The quantitative verification performance is presented in this section. The overall results demonstrate that our framework achieves state-of-the-art verification accuracy across multi-person status.
\subsection{COCO keypoint Detection}

\textbf{Dataset}

Our models are trained on the MSCOCO \cite{lin2014microsoft} train dataset (includes 57K images and 150K person instances with 17 keypoints) and validated on MSCOCO val dataset (includes 5k images). The testing sets include test-dev sets (20K images)

\textbf{Evaluation Method}

We report standard average precision(AP) and recall scores(AR): AP (the mean of AP scores at 10 positions,
OKS = 0.50; 0.55;...; 0.90; 095 $AP^{50}$ (OKS = 0.50) $AP^{75}$ (OKS = 0.75), ; $AP^M$ for medium
objects, $AP^L$ for large objects. Object Keypoint Similarity (OKS):
\begin{equation}
    OKS = \frac{{\sum\limits_{}^{} {{}_i{exp{( - \mathop d\nolimits_i^2 /2\mathop s\nolimits^2 \mathop k\nolimits_i^2 )\delta ({v_i} > 0)}}} }}{{\sum\limits_{}^{} {{}_i\delta ({v_i} > 0)} }}.
\end{equation}
$D_{pi}$ is the Euclidean distance between the predicted keypoints and ground-truth, $S_p$ is the scale factor, and the square root of the human body area. $\delta (\cdot)$ is included in the evaluation for the keypoints marked as visible (V=1). 

\textbf{Training Detials}

For each bounding-box, we crop the box from the image, which is resized to a fixed size, 384$\times$288. The we adopt data augmentation sequences previously searched by us. In order to get the sequence, we train from scratch with a global batch size 32, 640$\times$640 of images size, learning rate of 0.08, weight decay of 1e - 4, $\alpha$ = 0.25 and $\gamma$ = 1.5 for the focal loss parameters. 

We train for 100 epochs, using stepwise decay where the learning rate is reduced by a factor of 10 at epochs 70 and 90. All models are trained on 4 GPUs over 5 days. Meanwhile, in order to save computing resources and speed up, we only randomly use 5k COCO train images when searching for data augmentation sequences. The reward signal for the controller is the mAP on MS COCO val sets of 5,000 images. 

The models of pose estimation are trained using \textit{Adam} algorithm with an initial learning rate of 5$e^{-4}$. Note that we also decrease the learning rate by a factor of 2 every $3.6\times10^{6}$ iterations. We use a weight decay of 1$e^{-5}$ and set the training batch size 32. Batch normalization is used in our network. Generally, the training of ResNet-101 models takes about 1.5 days on 4 GPUs. Our models are initialized with weights of the ImageNet-pretrained model.

During the process on training, ParallelNet adopt L2 loss of all keypoints, at the same time RefineNet adopt $L2^*$ loss, which we select hard keypoints online based on traing loss(pixel-level heatmap L2 loss). We only keep the top $\alpha$($\alpha$=10) keypoints out of all \textit{N} keypoints.

\textbf{Testing Detials}

We apply a gaussian filter on the predicted
heatmaps, computing the heatmap by averaging the heatmaps of the original and flipped images. Each keypoint position is predicted by adjusting the highest calorific value position, which is shifted by a quarter in the direction from the highest response to the second highest response. We consider the product of boxes’ score and the average score of all keypoints as the final pose score of a person instance.
\begin{table*}[!t]
\caption{ Comparisons on COCO \textit{test-dev}.\#Params and FLOPs are calculated for the pose estimation network.}
\renewcommand{\arraystretch}{0.8}
\centering
    \setlength{\tabcolsep}{0.2mm}{
    \scriptsize
    \begin{tabular}{c|c|c|c|c|c|ccccc}
    \hline
    Method&Backbone&Input size&\#Params&GFLOPs&AP&$AP^{50}$&$AP^{75}$&$AP^{M}$&$AP^{L}$&AR\\
    \hline
    Mask-RCNN \cite{he2017mask}&ResNet-50-FPN&-&-&-&63.1&87.3&68.7&57.8&71.4&-\\
    G-RMI \cite{papandreou2017towards}&ResNet-101& 353$\times$257&42.6M& 57.0 &64.9&85.5& 71.3& 62.3 &70.0&69.7\\
    %G-RMI\cite{papandreou2017towards} + extra data&ResNet-101& 353$\times$257& 42.6M &  57.0 &68.5 &87.1 &  75.5 & 65.8 &73.3 &73.3\\
    CPN \cite{chen2018cascaded}&ResNet-Inception& 384$\times$288& - &  - &72.1&91.4 &  80.0 & 68.7&77.2&78.5\\
    RMPE \cite{fang2017rmpe}&PyraNet& 320$\times$256& 28.1M&  26.7 &72.3&89.2 &  79.1 & 68.0&78.6&-\\
    CFN \cite{huang2017coarse}&-& -& -&-&72.6&86.1 &  69.7 & 78.3&64.1&-\\ 
    CPN \cite{chen2018cascaded} (ensemble)&ResNet-Inception& 384$\times$288& - &  - &73.0&91.7 &  80.9 & 69.5&78.1&79.0\\
    SimpleBaseline \cite{xiao2018simple}&ResNet-152& 384$\times$288& 68.6M &  35.6  &73.7&91.9&  81.1 & 70.3&80.0&79.0\\
    HRNet-W32 \cite{sun2019deep} & HRNet-W32 & 384$\times$288& 28.5M &  16.0  &74.9& 92.5 &  82.8 & 71.3 & 80.9 &80.1\\
    \textbf{Ours}&\textbf{ResNet101}&\textbf{384}\textbf{$\times$}\textbf{288}& \textbf{47.5M} &\textbf{-}&\textbf{76.5}&\textbf{92.1}&  \textbf{83.7}&\textbf{73.2}&\textbf{82.2}&\textbf{82.5}\\
   
    \hline
    \end{tabular}
    \label{tab.2}
    }
\end{table*}

\begin{table*}[!t]
    \caption{Performance comparisons on the MPII test set (PCKh@0.5).}
    \renewcommand{\arraystretch}{1.0}
    \centering
    \begin{tabular}{l|ccccccc|c}
    \hline
    Method & Hea&Sho&Elb&Wri&Hip&Kne&Ank&Total\\
    \hline
   Stack Hourglass \cite{newell2016stacked}. &98.2 &96.3& 91.2&87.1&90.1&87.4&83.6&90.9\\
   Sun et al \cite{sun2017human}. &98.1 &96.2& 91.2&87.2&89.8&87.4&84.1&91.0\\
   Chu et al \cite{chu2017multi}. &98.5 &96.3& 91.9&88.1&90.6&88.0&85.0&91.5\\
   Chou et al \cite{chou2018self}. &98.2 &96.8& 92.2&88.0&91.3&89.1&84.9&91.8\\
   Yang et al \cite{yang2017learning}. &98.5 &96.7& 92.5&88.7&91.1&88.6&86.0&92.0\\
   Ke et al \cite{ke2018multi}. &98.5 &96.8& 92.7&88.4&90.6&89.3&86.3&92.1\\
   Tang et al \cite{tang2018deeply}. &98.4 &96.9& 92.6&88.7&91.8&89.4&86.2&92.3\\
   SimpleBaseline \cite{xiao2018simple}&98.8 &96.6& 91.9&87.6&91.1&88.1&84.1&91.5\\
   HRNet-W32 \cite{sun2019deep}&98.6 &96.9& 92.8&89.0&91.5&89.0&85.7&92.3\\
   \textbf{Ours}&\textbf{98.7} &\textbf{97.1}& \textbf{92.9}&\textbf{89.2}&90.1&\textbf{90.5}&\textbf{85.8}&\textbf{92.4}\\
    \hline
    \end{tabular}
    \label{tab.3}

\end{table*}

\subsection{MPII keypoint Detection}

\textbf{Dataset}

There are 25K images with 40K instances, where there are 12K subjects for testing and the remaining subjects for the training set. The data augmentation and the training strategy are the same to MS COCO. The standard metric, the PCKh (head-normalized probability of correct keypoint) score, is
used. The PCKh@0:5 ($\alpha$ = 0.5) score is reported.

\subsection{Experiment Results}

Tables~\ref{tab.2},\ref{tab.3} show the pose-estimation performance of our method. We get state-of-the-art results on both two datasets, MS COCO test-dev and MPII. Our performance is better than SimpleBaseline, but the amount of parameters is much less than SimpleBaseline.

\subsection{Ablation Study} 
\textbf{Components of $P^2$ Net}. We analyze the important of each component in $P^2$ Net. Our net is evaluated on ResNet101 backbone.

\begin{itemize}
    \item \textbf{Detector Network}. Table~\ref{tab.4} shows the relationship between detection AP and the corresponding keypoints AP, we have chosen the three detector networks, Faster R-CNN \cite{ren2015faster}, DetNet \cite{li2018detnet} and TridentNet \cite{li2019scale}. From the table, when the detection AP increases and the human detection AP does not increases. Therefore, the more important task for pose-estimation is to enhance the accuracy of the keypoints rather than involve more boxes.
     \begin{table}[htb]
    \caption{ Comparing with different detector networks. $AP^*$ means the performance of $P^2$ Net based on the specific detector network. }
    \renewcommand{\arraystretch}{1.2}
    \centering
        \setlength{\tabcolsep}{0.3mm}{
        \begin{tabular}{c|c|c|c|c|c|c|c}
        \hline
        Detector&Backbone&mAP&$AP^{50}$&$AP^{S}$&$
        AP^{M}$&$AP^{L}$&$AP^*$\\
        \hline
        Faster R-CNN&ResNet50&37.4&59.0&18.3&41.7&52.9&\textbf{76.2}\\
        DetNet59&ResNet50&40.2&61.7&\textbf{23.9}&43.2&52.0&\textbf{76.3}\\
        TridentNet&ResNet101&\textbf{40.6}&\textbf{61.8}&23.0&\textbf{45.5}&\textbf{55.9}&\textbf{76.3}\\
        \hline
        \end{tabular}
        \label{tab.4}
        }
    \end{table}

    \item \textbf{Attention Partial Module}. As shown in Table~\ref{tab.5},comparing our approach to the method based on our network without APM, we can see that performance has improved a lot. This indicates that APM can gather features more selectivity and efficiently from shallow layers and deep layers.
    \begin{table}[htb]
    \caption{ Compared with the result without Attention Module. $Ours^*$ means the method without Attention Module, adopting the auto data augmentations. }
    \renewcommand{\arraystretch}{1.2}
    \centering
        \setlength{\tabcolsep}{0.6mm}{
        \begin{tabular}{c|c|c|c|c|c|c|c}
        \hline
        Method&Backbone&AP&$AP^{50}$&$AP^{75}$&$
        AP^{M}$&$AP^{L}$&AR\\
        \hline
        \textbf{Ours}&\textbf{ResNet101}&\textbf{76.5}&\textbf{92.1}&  \textbf{83.7}&\textbf{73.2}&\textbf{82.2}&\textbf{82.5}\\
        $Ours^*$&ResNet101&76.2&92.5&83.1&72.2&82.2&81.4\\
        \hline
        \end{tabular}
        \label{tab.5}
        }
    \end{table}
    
\end{itemize}

\section{Conclusion}
This paper has proposed a top-down pose-estimation framework with auto data augmentations. In human detection stage of our framework, Det-Net has been employed to enhance the coherence between the ImageNet pretrained model and detection model, resulting in better representation ability of feature maps. The detected human images are then augmented by searched policies. A differentiable method is to search for different augmentations, policies, which are regarded as different routines to be selected. Finally, the ParallelNet is designed to fuse feature maps of different levels, such that both high- and low-level information can be utilized. Our proposed framework pays attention to all stages in pose estimation, therefore can significantly reduce estimation errors. The experimental results demonstrate that our proposed framework outperforms state-of-the-art human pose-estimation methods.
% ---- Bibliography ----
%
% BibTeX users should specify bibliography style 'splncs04'.
% References will then be sorted and formatted in the correct style.
%
\bibliographystyle{splncs04}
\bibliography{egbib}
\end{document}